\documentclass[10pt,a4paper,reqno]{amsart}
\usepackage{amsthm}
\usepackage{amsmath}
\usepackage{amsmath, amssymb}
\usepackage{type1cm}
\usepackage{cite}
\usepackage{bm}
\usepackage{cases}
\usepackage{here}
\usepackage[pdftex]{graphicx}
\usepackage{remreset}
\makeatletter
	
	\@addtoreset{figure}{section}

	\@addtoreset{table}{section}
\makeatother
\theoremstyle{definition}
\newtheorem{defi}{Definition}[section]

\newcommand{\eqdef}{\stackrel{\mathrm{def}}{=}}
\begin{document} 
\title{Divergence Network: Graphical calculation method of divergence functions}
\author{Tomohiro Nishiyama}
\begin{abstract}
In this paper, we introduce directed networks called ``divergence network'' in order to perform graphical calculation of divergence functions.
By using the divergence networks, we can easily understand the geometric meaning of calculation results and grasp relations among divergence functions intuitively. 
\\

\end{abstract}
\date{}
\maketitle
\bibliographystyle{plain}
\section{Introduction}
Divergences are functions which measure the discrepancy between two points and play a key role in the field of machine learning, statistics, signal processing.
Given a set $\Omega$ and $P,Q\in \Omega$, a divergence is defined as a function $D:\Omega\times \Omega\rightarrow \mathbb{R}$ which satisfies the following properties.\\
1. $D(P,Q)\geq 0 $ for all $P,Q\in  \Omega$.\\
2. $D(P,Q)=0\iff P=Q$.\\

When analyzing the relations among multiple classes of divergence functions or analyzing divergence functions for multiple points,  we must deal with complicated calculation and it is not easy to understand the geometric meaning of the results.

Therefore, we introduce new directed networks called ``divergence network'' for graphical calculation of divergence functions.

In section 2, we define the divergence networks and the network function.

In section 3, we show properties of the network function and the deformation rules of the divergence networks.

Section 2 and 3 are the main contents of this paper. 

In section 4, we show network representations of some divergence functions (the Bregman divergence\cite{bregman1967relaxation}, the Jensen divergence\cite{nielsen2011burbea, burbea1980convexity}, $f$-divergence\cite{csiszar1967information,ali1966general} and so on) and in section 5 we derive some results shown in \cite{nishiyama2018generalized ,nishiyama2018sum, nielsen2009sided} by using the divergence networks. These classes of divergences include well-known divergence functions (e.g. the Kullback-Leibler divergence\cite{kullback1997information}, the Jensen-Shannon divergence\cite{burbea1980convexity} and so on).

\section{Divergence Network and network function}
\label{divergence_network}
First, we define the convex conjugate.
\begin{defi}
The convex conjugate $F^\ast :\mathrm{dom} F^\ast\rightarrow\mathbb{R}$ is defined in terms of the supremum by
\begin{align}
F^\ast(x^\ast)\eqdef \sup_x \{\langle x^\ast, x\rangle - F(x)\},
\end{align}
where $\langle\cdot, \cdot\rangle$ is an inner product.
$F^\ast$ is a convex function.
\end{defi}
Because the derivative of $F$ is the maximizing argument, we get
\begin{align}
\label{convex conjugate}
x^\ast=\nabla F(x) \\ \nonumber
F(x)+F^\ast(x^\ast)=\langle x^\ast, x\rangle.
\end{align}

\subsection{Notations and definitions}
\begin{itemize}
\item Let $i,j \in\mathbb{N}$ be an indices.
\item Let $P_i, Q_i$ be coordinates in $\mathbb{R}^d$.
\item Let $\alpha_i, \beta_i$ be weights in $\mathbb{R}$.
\item Let  $\langle\cdot, \cdot\rangle$ be an inner product.
\item Let $F:\mathbb{R}^d \rightarrow\mathbb{R}$ be a differentiable and strictly convex function.
\item Let $F^\ast$ be a convex conjugate function of $F$.
\item Let $x^\ast=\nabla F(x)$.
\item $F(P)$ denotes $F(P_1, P_2, \cdots, P_d)$.

\end{itemize} 

\begin{defi}(Definition of the divergence network)
Let $\{v_i\}$ be nodes with coordinates and ON or OFF states.

Let $\{e_j\}$ be directed or undirected edges with weights and ON or OFF states.

Let each edge be associated with two nodes.

Let $V$ be a set of nodes $\{v_i\}$ and $E$ be a set of edges $\{e_j\}$.

We define the divergence network $N$ as an ordered pair $N\eqdef (V, E)$.
\end{defi}

\begin{defi}(Definition of the network function)
\label{def_netfunc} 
Let $\Lambda$ be a set of the divergence networks.
Let $N,N_1\in\Lambda$.
Let $V(N)$ be all nodes in $N$ and let $E(N)$ be all edges in $N$.
We define the network function $\Phi_N :\Lambda\rightarrow \mathbb{R}$ as follows.
\begin{align}
\Phi_N(N_1)\eqdef \sum_{V(N_1)} \Phi_N(v_i) + \sum_{E(N_1)} \Phi_N(e_j)
\end{align}
We take the sum of all nodes and edges in the divergence network $N_1$.
\end{defi}
When $N=N_1$, we omit the subscription $N$ of $\Phi_N$ and write $\Phi_N(N)$ as $\Phi(N)$.

When $v_i, e_j\in N$, we define the value of $\Phi_N(v_i)$ and $\Phi_N(e_j)$ in the following subsections.

When $v_i, e_j\notin N$, we define $\Phi_N(v_i)=\Phi_N(e_j)=0$.\\
\noindent\textbf{Remark:} However the network function $\Phi$ also depends on the convex function $F$, we don't explicitly write $F$ since we deal with only a single convex function in this paper.

\begin{defi}(Symbols about the divergence networks)
Let $N, N_1, N_2 $ be the divergence networks.
\begin{itemize}
\item Let $V(N)$ be all nodes in $N$.
\item Let $E(N)$ be all edges in $N$.
\item We write $N_1=N_2$ if and only if $V(N_1)=V(N_2)$ and $E(N_1)=E(N_2)$.
\item When $V(N)=V(N_1)\cup V(N_2)$ and $E(N)=E(N_1)\cup E(N_2)$, we write $N=N_1\cup N_2$.\\
\item When $V(N)=V(N_1)\cap V(N_2)$ and $E(N)=E(N_1)\cap E(N_2)$, we write $N=N_1\cap N_2$.\\
\item When  $N=N_1\cup N_2$ and $E(N_1)\cap E(N_2)=\phi$, we write $N=N_1+N_2$.
\item When $\Phi_N(N_1)=\Phi_N(N_2)$, we write $N_1\iff N_2$.
This symbol means that we can deform $N_1$ to $N_2$ and  $N_2$ to $N_1$ while keeping the value of the network function.
\end{itemize}
\end{defi}

\subsection{Edges}
We introduce directed edges (arrows).
\begin{figure}[H]
 \begin{center}
  \includegraphics[width=120mm]{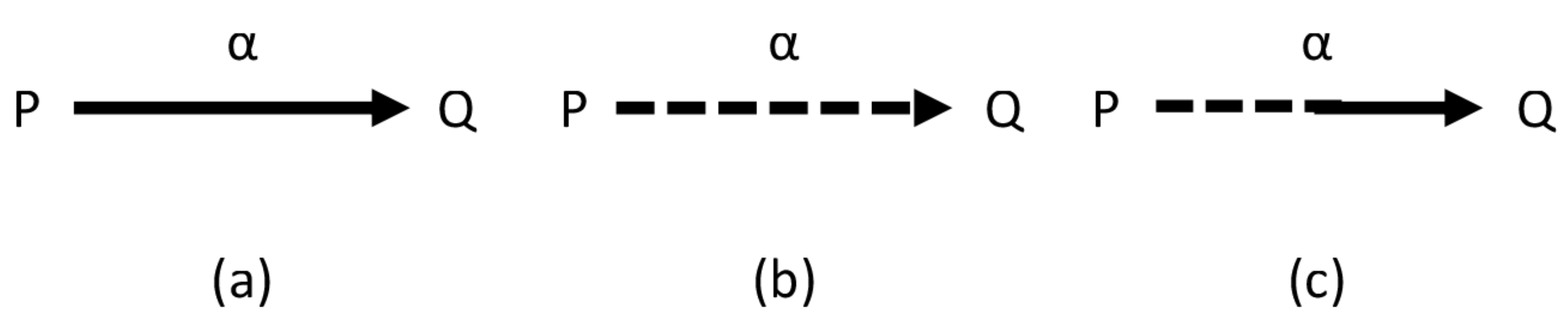}
 \end{center}
 \caption{Definition of arrows.}
 \label{def_arrows}
\end{figure}
$\alpha\in\mathbb{R}$ denotes a weight and $P,Q\in\mathbb{R}^d$ denote coordinates.
\begin{itemize}
\item (a) ON-arrow. \\
\begin{align}
\Phi_N(e)\eqdef -\alpha\langle P, Q^\ast \rangle
\end{align}
\item (b) OFF-arrow. \\
\begin{align}
\Phi_N(e)\eqdef 0
\end{align}

\item (c) denotes ON or OFF-arrow. \\
\end{itemize}
\textbf{Remark:}
We regard ON-arrow and OFF-arrow which connect $P$ and $Q$ as different edges.
If $P=Q$, an arrow becomes a directed loop. 
Next, we introduce undirected edges (lines).
\begin{figure}[H]
 \begin{center}
  \includegraphics[width=120mm]{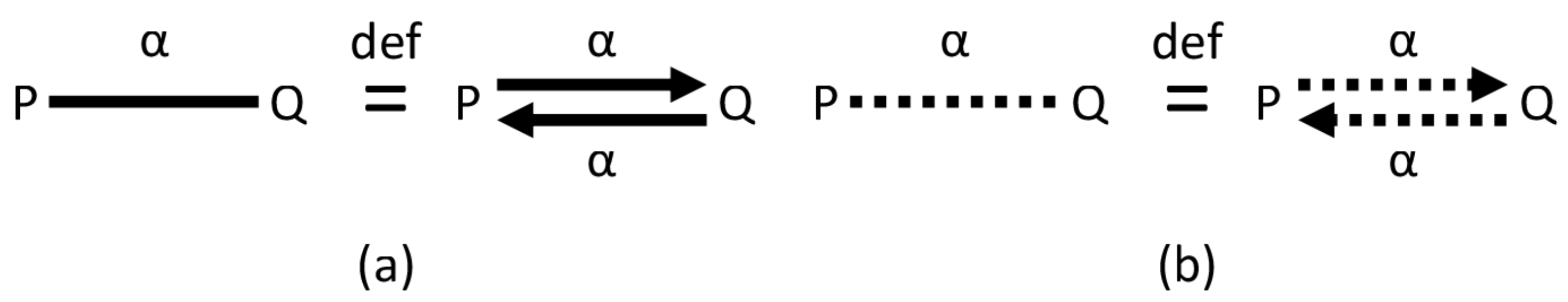}
 \end{center}
 \caption{Definition of lines.}
 \label{def_lines}
\end{figure}
If $P=Q$, a line becomes a undirected loop. 
\subsection{Nodes}
We introduce nodes.
\begin{figure}[H]
 \begin{center}
  \includegraphics[width=50mm]{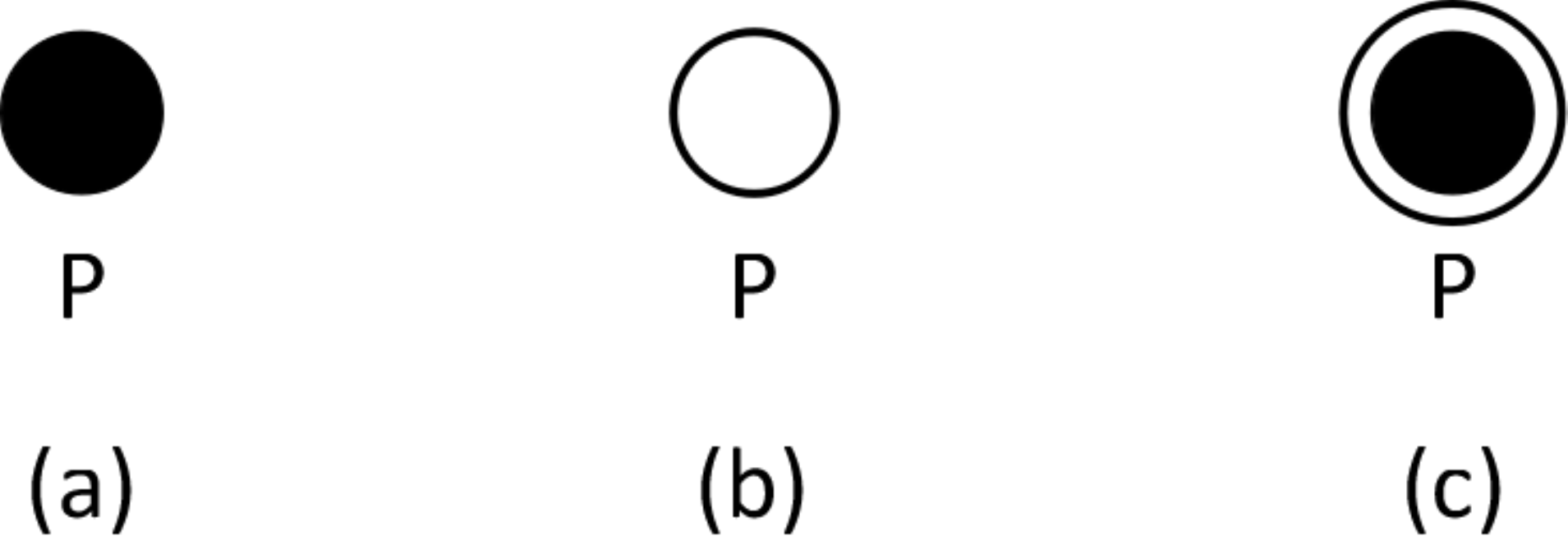}
 \end{center}
 \caption{Definition of nodes.}
 \label{def_nodes}
\end{figure}
$P\in\mathbb{R}^d$ denotes a coordinate of a node.
\begin{itemize}
\item (a) ON-node. \\
\begin{align}
\Phi_N(v)\eqdef \sum_i\alpha_i F^\ast(P^\ast) + \sum_j\beta_j F(P),
\end{align} 
where $\sum_i\alpha_i $ denotes a total sum of input weights of both solid and dotted edges and $\sum_j\beta_j $ denotes a total sum of output weights of both solid and dotted edges. If $\mathrm{deg}(v)=0$, $\Phi_N(v)=0$.
\item (b) OFF-node. \\
\begin{align}
\Phi_N(v)\eqdef 0
\end{align}
\item (c) denotes ON or OFF-node. \\
\end{itemize}
\textbf{Remark:}
Even if the coordinates of nodes are the same, we regard ON-node and OFF-node as different nodes.

Next, we introduce a centroid and convex conjugate centroid.
These centroids are a kind of nodes.

\begin{figure}[H]
 \begin{center}
  \includegraphics[width=40mm]{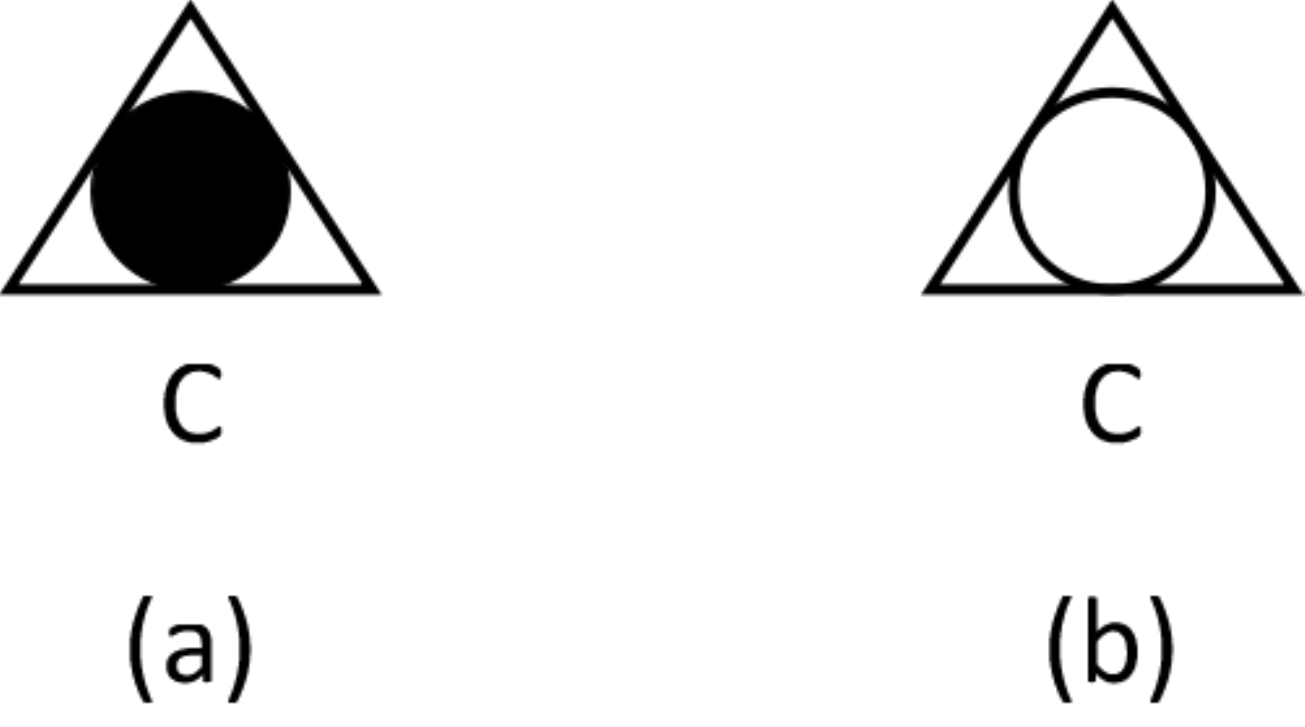}
 \end{center}
 \caption{Definition of centroids.}
 \label{def_centroids}
\end{figure}
(a) denotes ON-centroid and (b) denotes OFF-centroid.
The values of the network function for centroids are defined the same as nodes.
A coordinate $C$ is defined as 
\begin{align}
C\eqdef \frac{1}{\Sigma}\sum_i \alpha_i P_i,
\end{align}
 where $\Sigma\eqdef \sum_i \alpha_i$, $\{\alpha_i\}$ are the weights of solid or dotted arrows and $\{P_i\}$ are \textbf{the start points} of solid or dotted arrows. 

\begin{figure}[H]
 \begin{center}
  \includegraphics[width=40mm]{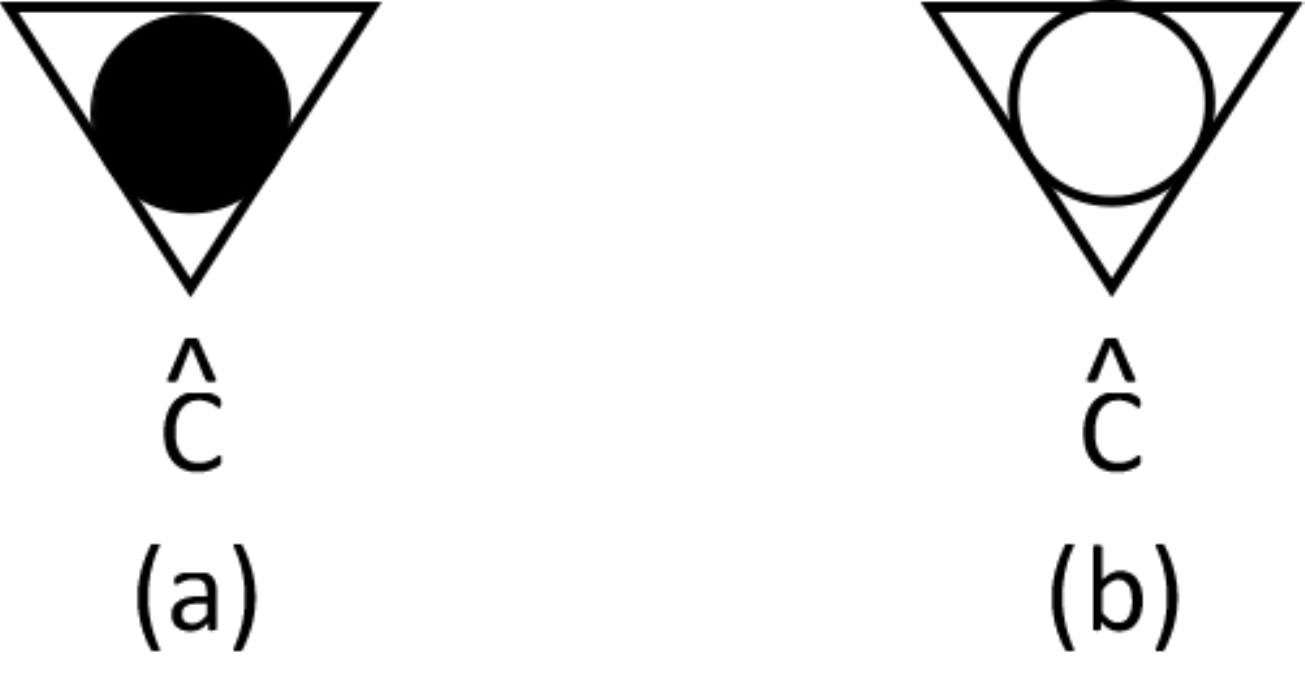}
 \end{center}
 \caption{Definition of convex conjugate centroids.}
 \label{def_conjugate_centroids}
\end{figure}
(a) denotes ON-convex conjugate centroid and (b) denotes OFF-convex conjugate centroid.
The values of the network function for convex conjugate centroids are defined the same as nodes.
A coordinate $\hat{C}$ is defined as 
\begin{align}
\hat{C}^\ast\eqdef \frac{1}{\Sigma}\sum_i \alpha_i P_i^\ast,
\end{align}
where $\Sigma\eqdef \sum_i \alpha_i$, $\{\alpha_i\}$ are the weights of solid or dotted arrows and $\{P_i\}$ are \textbf{the end points} of solid or dotted arrows. 

\section{Properties of the network function and Deformation rules}
First, we introduce a simple but important divergence network called ``Bregman network''.

Next, we show properties of the network function.

Finally, we introduce deformation rules for the divergence networks.
\subsection{Bregman network}
\label{Bregman_network}
We define the Bregman network $N_B$ as follows.
\begin{figure}[H]
 \begin{center}
  \includegraphics[width=30mm]{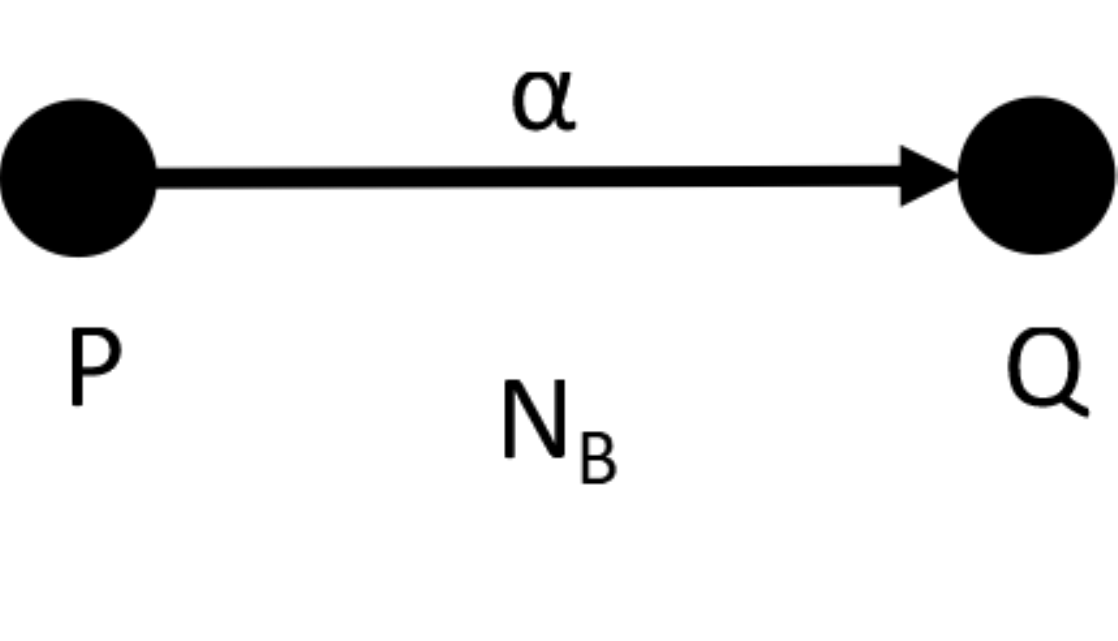}
 \end{center}
 \caption{Bregman network.}
 \label{Bregman}
\end{figure}
\textbf{Properties of the Bregman network}
\begin{itemize}
\item $\Phi(N_B)=\alpha B_F(P,Q)$, where $B_{F}(P,Q)\eqdef F(P)-F(Q)-\langle \nabla F(Q), P-Q\rangle$ is the Bregman divergence. 
\item If $\alpha > 0$, $\Phi(N_B)\geq 0$ and $\Phi(N_B)=0\iff P=Q$.
\item If $\alpha < 0$, $\Phi(N_B)\leq 0$ and $\Phi(N_B)=0\iff P=Q$.
\end{itemize}
\noindent \textbf{Proof.}\\
From Figure \ref{def_arrows} and Figure \ref{def_nodes} and using (\ref{convex conjugate}), we get
\begin{align}
\label{eq_Bregman}
\Phi(N_B)&=\alpha(F(P)-\langle P, Q^\ast \rangle + F^\ast(Q^\ast))=\alpha(F(P)-\langle P, Q^\ast \rangle +\langle Q, Q^\ast \rangle - F(Q))\\ \nonumber
&=\alpha(F(P)-F(Q)-\langle Q^\ast, P-Q\rangle)=\alpha B_F(P,Q).
\end{align}
About other properties, we can easily verify by using properties of the Bregman divergence.

\subsection{Properties of the network function}
Let $N, N_1, N_2$ be the divergence networks.\\
\textbf{Property 1.}\\
If $N=N_1+N_2$, 
\begin{align}
\Phi(N)=\Phi(N_1)+\Phi(N_2).
\end{align}
\textbf{Property 2.}\\
Let $\{N_{B,i}\}$ be the Bregman networks and let $N=\sum_i N_{B,i}$.
Let all weights in $N$ are positive.
Then, $\Phi(N)$ is a divergence function.
\begin{itemize}
\item $\Phi(N)\geq 0$.
\item $\Phi(N)=0$ if and only if coordinates of all nodes in $N$ are the same.
\end{itemize}
\textbf{Property 3.}
Let $N_B$ be the Bregman network. 
Let $N=N_1+N_B$ and let $\alpha$ be a weight of the Bregman Network.
If $\alpha > 0$,
\begin{align}
\Phi(N) \geq \Phi(N_1).
\end{align}
If $\alpha < 0$,
\begin{align}
\Phi(N) \leq \Phi(N_1).
\end{align}
\noindent\textbf{Proof of Property 1.}\\
From $E(N_1)\cap E(N_2)=\phi$ and the definition of $\Phi$ for edges, we get
$\Phi_N(e_j)=\Phi_{N_1}(e_j)+\Phi_{N_2}(e_j)$.

For nodes, we first consider the case $v_i\in V(N)\setminus V(N_2)$.
By the definition of the divergence network, all edges which are associated with the node $v_i$ belong to $N_1$.
From the definition of $\Phi$ for nodes, we get $\Phi_N(v_i)=\Phi_{N_1}(v_i)$ and $\Phi_{N_2}(v_i)=0$.
Hence, we get  $\Phi_N(v_i)=\Phi_{N_1}(v_i)+\Phi_{N_2}(v_i)$.

In the case  $v_i\in V(N)\setminus V(N_1)$, we get $\Phi_N(v_i)=\Phi_{N_1}(v_i)+\Phi_{N_2}(v_i)$ in the same way.

Finally, we consider the case $v_i\in V(N_1)\cap V(N_2)$. From $E(N_1)\cap E(N_2)=\phi$ and the definition of $\Phi$ for nodes, we get $\Phi_N(v_i)=\Phi_{N_1}(v_i)+\Phi_{N_2}(v_i)$.
From these result, we get 
\begin{align}
\Phi(N)&= \sum_{V(N)} \Phi_N(v_i) + \sum_{E(N)} \Phi_N(e_j)\\ \nonumber
&=\bigl(\sum_{V(N)} \Phi_{N_1}(v_i) + \sum_{E(N)} \Phi_{N_1}(e_j) \bigr)+ \bigl(\sum_{V(N)} \Phi_{N_2}(v_i) + \sum_{E(N)} \Phi_{N_2}(e_j)\bigr)\\ \nonumber
&=\bigl(\sum_{V(N_1)} \Phi_{N_1}(v_i) + \sum_{E(N_1)} \Phi_{N_1}(e_j) \bigr)+ \bigl(\sum_{V(N_2)} \Phi_{N_2}(v_i) + \sum_{E(N_2)} \Phi_{N_2}(e_j)\bigr)\\ \nonumber
&=\Phi(N_1)+\Phi(N_2).
\end{align}
\noindent\textbf{Proof of Property 2 and 3.}\\
By Property 1 and the properties of the Bregman network, we get the results.

Then, we introduce deformation rules of the divergence network.
\subsection{Summation rule}
We can integrate arrows of the same state which are associated with the same nodes.
\begin{figure}[H]
 \begin{center}
  \includegraphics[width=120mm]{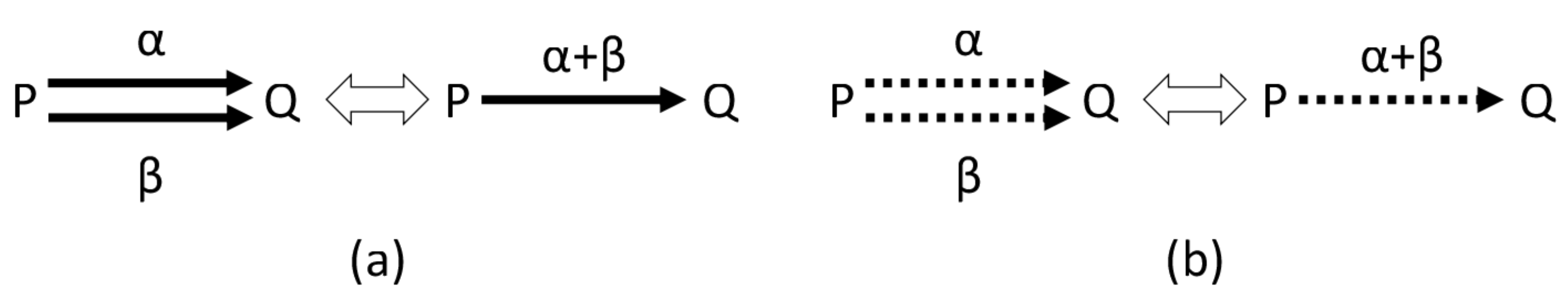}
 \end{center}
 \caption{Summation rule.}
 \label{sum_rule}
\end{figure}
We can easily prove by the definition of $\Phi$ for nodes and edges.

\subsection{Deletion rule}
We can delete an isolated node $v$ ($\mathrm{deg}(v)=0$).
We can easily verify by using the definition of $\Phi$ for nodes.

We can also delete all edges in the following figure.
\begin{figure}[H]
 \begin{center}  \includegraphics[width=80mm]{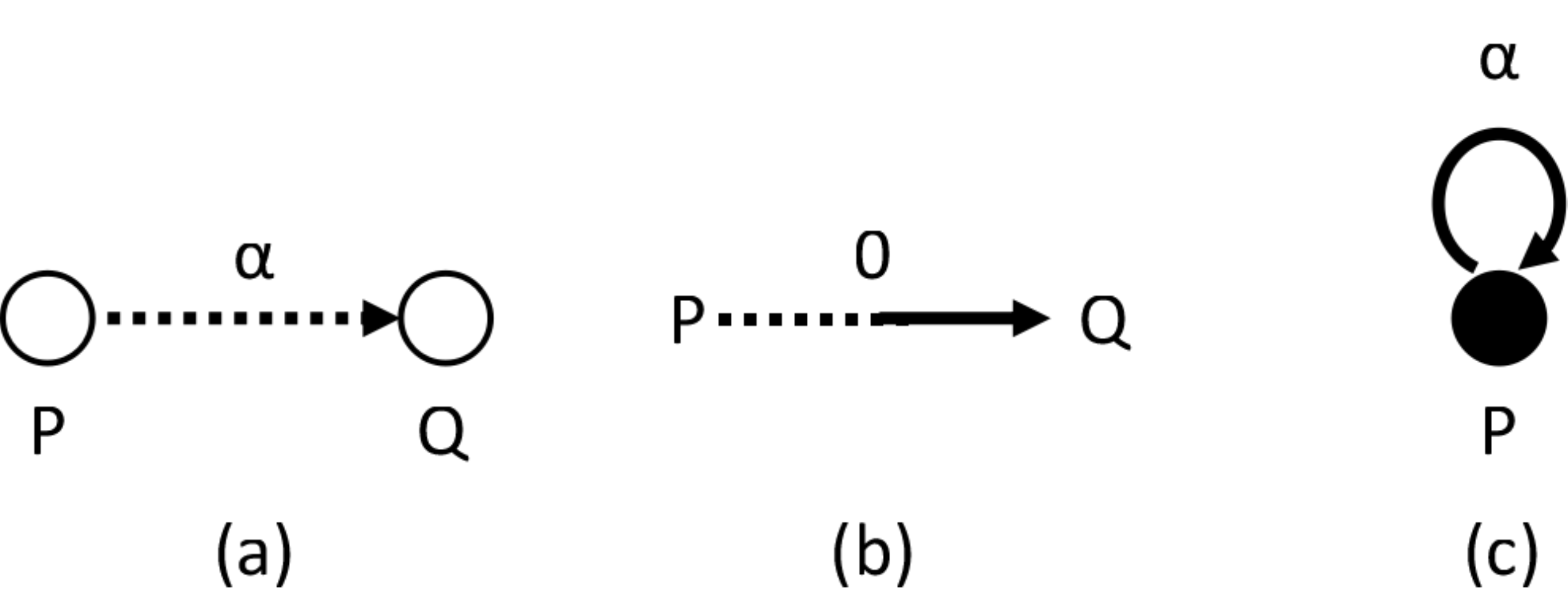}
 \end{center}
 \caption{Deletion rule.}
 \label{deletion_rule}
\end{figure}
We can easily verify for Figure \ref{deletion_rule} (a) and (b).
For Figure \ref{deletion_rule} (c), we can prove by putting $P=Q$ in Figure \ref{Bregman} and using $B_F(P,P)=0$.
\subsection{ON-OFF rule 1}
Let weights satisfy $\Sigma\eqdef \sum_{i=1}^M \alpha_i=\sum_{j=1}^L \beta_j$ and let $v$ be a node with a coordinate $P$.
In the following case, we can change the state of node $v$.
\begin{figure}[H]
 \begin{center}
  \includegraphics[width=100mm]{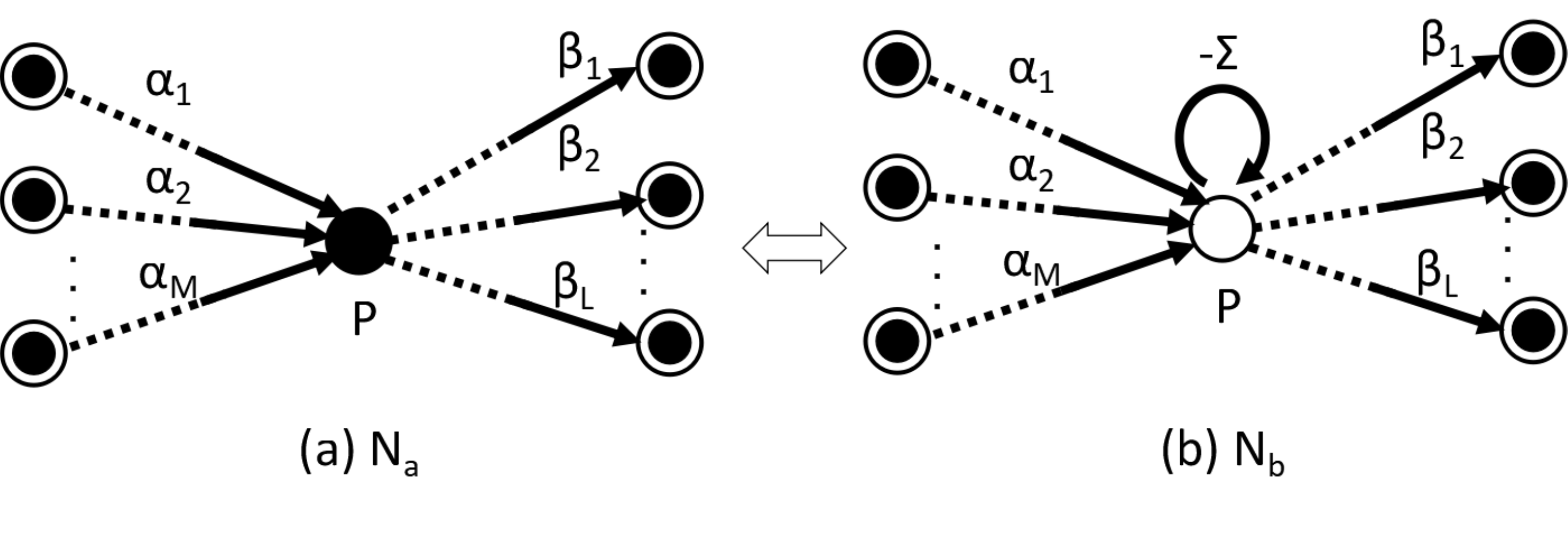}
 \end{center}
 \caption{ON-OFF rule 1.}
 \label{onoff_rule1}
\end{figure}
\noindent\textbf{Remark:} $N_a$ and $N_b$ are the same other than the state of $v$ and a loop on $v$.\\
\noindent\textbf{Proof.}\\
From the definition of the network function for nodes and edges and using (\ref{convex conjugate}), we get
\begin{align}
\Phi_N(N_a)&=\sum_i \alpha_i F^\ast(P^\ast) + \sum_i \beta_i F(P)+R(v,e)=\Sigma \langle P,P^\ast\rangle + R(v,e)\\
\Phi_N(N_b)&=\Sigma \langle P,P^\ast\rangle +R(v,e), 
\end{align}
where $R(v,e)$ denotes the sum of the values from nodes other than $v$ and edges other than the loop on $v$.
Hence, $\Phi_N(N_a)=\Phi_N(N_b)$.

\subsection{ON-OFF rule2}
Let weights satisfy $\Sigma\eqdef \sum_{i=1}^M \alpha_i=\sum_{j=1}^L \beta_j$ and let $v$ be a node with a coordinate $P$.
In the following case, we can change the state of node $v$.
\begin{figure}[H]
 \begin{center}
  \includegraphics[width=100mm]{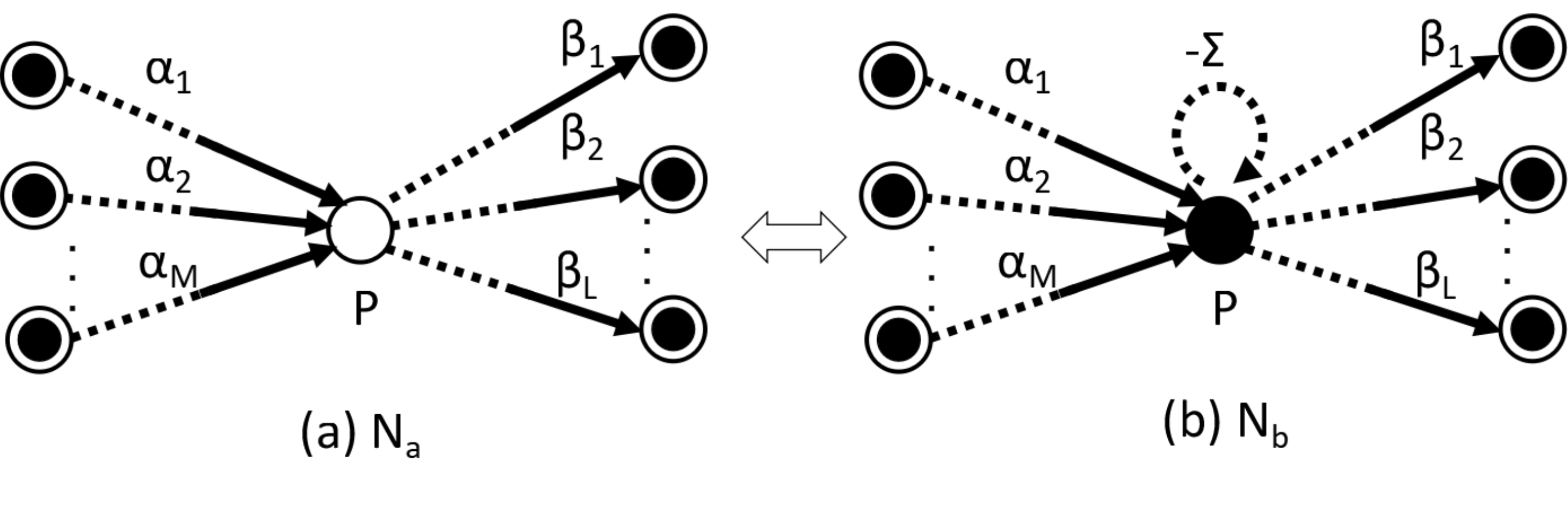}
 \end{center}
 \caption{ON-OFF rule 2.}
 \label{onoff_rule2}
\end{figure}
\noindent\textbf{Remark:} $N_a$ and $N_b$ are the same other than the state of a node $v$ and a directed loop on $v$.\\
\noindent\textbf{Proof.}\\
From the definition of the network function for nodes and edges, we get
\begin{align}
\Phi_N(N_a)&=R(v,e)\\
\Phi_N(N_b)&=\sum_i \alpha_i F^\ast(P^\ast) + \sum_i \beta_i F(P)-\Sigma(F(P) + F^\ast(P^\ast) ) +R(v,e)=R(v,e), 
\end{align}
where $R(v,e)$ denotes the sum of the values from nodes other than $v$ and edges other than the loop on $v$.
Hence, $\Phi_N(N_a)=\Phi_N(N_b)$.

\subsection{Insertion rule 1}
Let $\Sigma\eqdef \sum_{i=1}^M \alpha_i$ and $C\eqdef \frac{1}{\Sigma}\sum_{i=1}^M \alpha_i P_i$. Let $v$ be a node with a coordinate $Q$.
In the following case, we can insert a centroid.
\begin{figure}[H]
 \begin{center}
  \includegraphics[width=120mm]{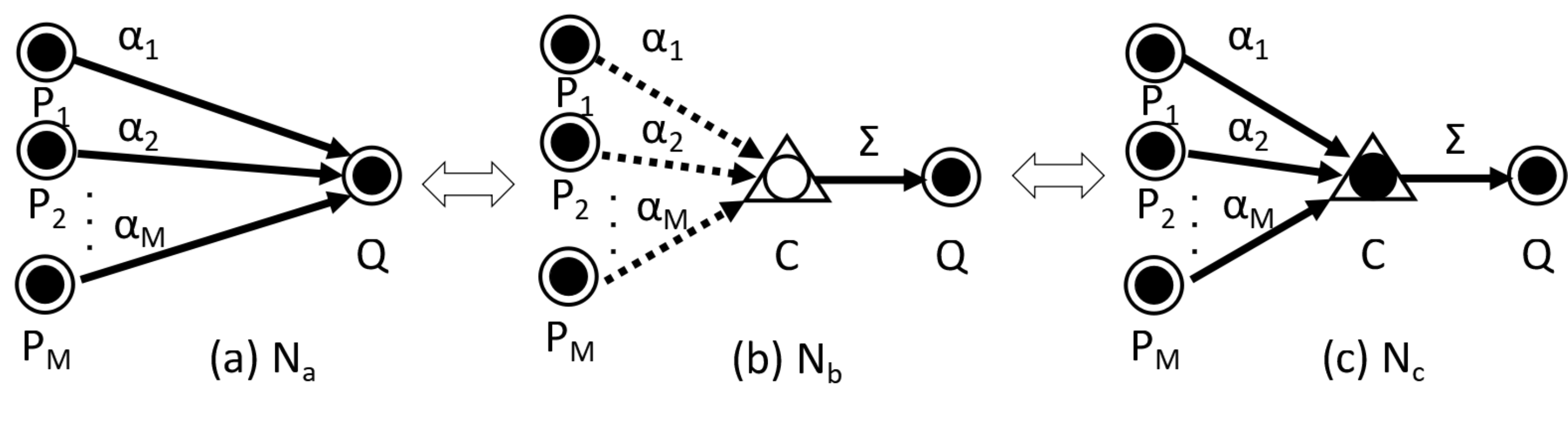}
 \end{center}
 \caption{Insertion rule 1.}
 \label{insertion_rule1}
\end{figure}
\noindent\textbf{Remark:} In $N_a$, $N_b$ and $N_c$, the state of nodes with coordinates $\{P_i\}$ and $Q$ are the same.\\
\noindent\textbf{Proof.}\\
We first prove Figure \ref{insertion_rule1} (a) $\rightarrow$ (b) in the case a node $Q$ is ON-state.
By using the definition of centroids $C\eqdef \frac{1}{\Sigma}\sum_i \alpha_i P_i$, we get 
\begin{align} 
\Phi_N(N_a)&= \sum_i\alpha_i\langle P_i, Q^\ast\rangle + \sum_i \alpha_i F^\ast(Q^\ast) + R(v)\\ \nonumber
&=\Sigma \langle C, Q^\ast\rangle + \Sigma  F^\ast(Q^\ast) + R(v)\\
\Phi_N(N_b)&=\Sigma \langle C, Q^\ast\rangle + \Sigma  F^\ast(Q^\ast) + R(v),
\end{align}
where $R(v)$ denotes the sum of the values from nodes with coordinates $\{P_i\}$.
Hence, the result follows. In the case a node $Q$ is OFF-state, we can prove in the same way.

Next, we prove Figure \ref{insertion_rule1} (b) $\iff$ (c) by using the deformation rules as follows.
\begin{figure}[H]
 \begin{center}
  \includegraphics[width=90mm]{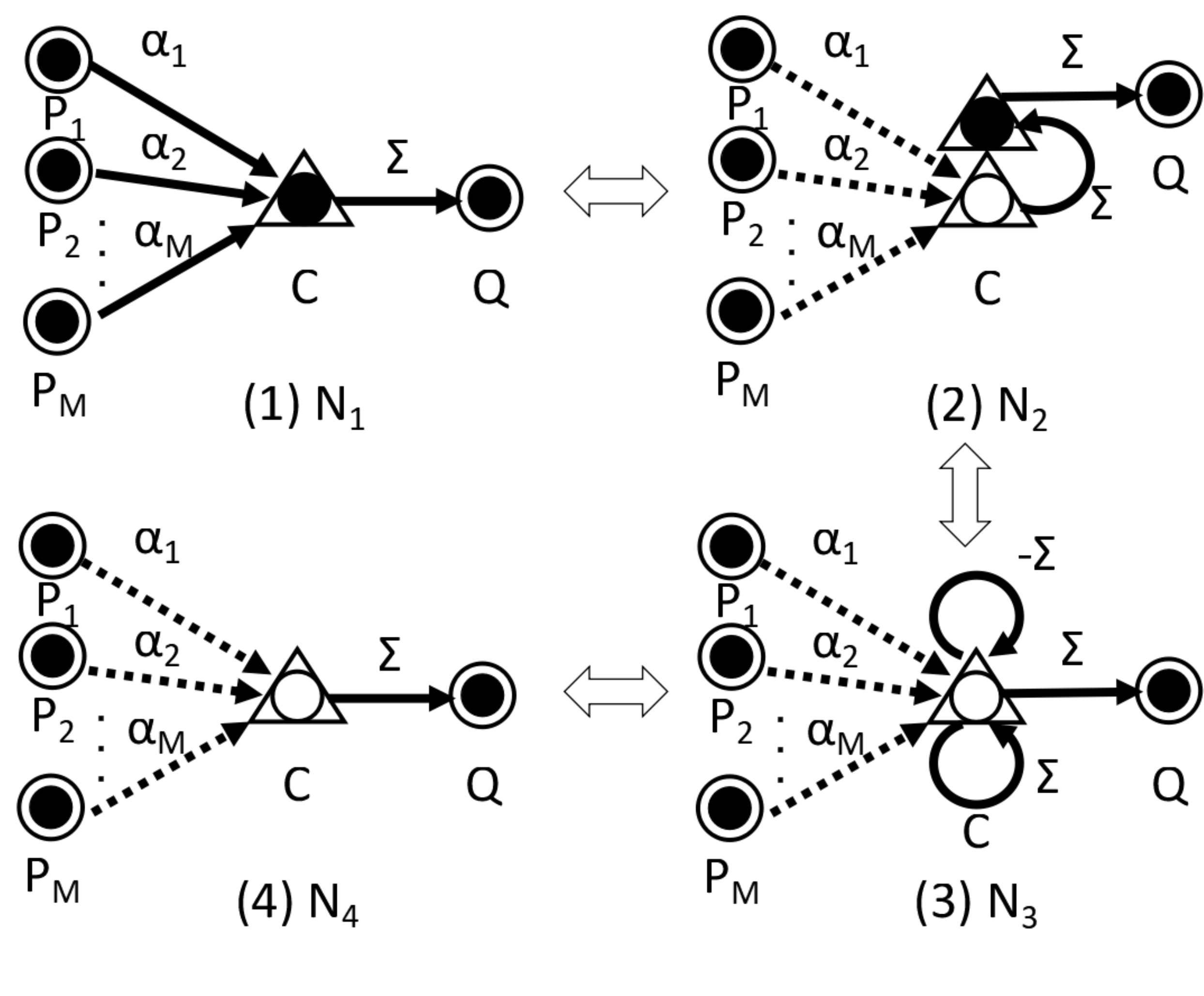}
 \end{center}
 \caption{Deformation of the network $N_c$.}
 \label{insertion_proof}
\end{figure}
\begin{itemize}
\item (1)$\rightarrow$(2): We apply Figure \ref{insertion_rule1} (a)$\rightarrow$ (b) (insertion rule 1) for $Q=C$.
\item (2)$\rightarrow$(3): We apply Figure \ref{onoff_rule1} (a)$\rightarrow$ (b) (ON-OFF rule 1) for $C$.
\item (3)$\rightarrow$(4): We apply Figure \ref{sum_rule} (a) (summation rule) and Figure \ref{deletion_rule} (b) (deletion rule).
\end{itemize}
Because $N_4=N_b$, the result follows

\subsection{Insertion rule2}
Let $\Sigma\eqdef \sum_{i=1}^M \alpha_i$ and $\hat{C}^\ast\eqdef \frac{1}{\Sigma}\sum_{i=1}^M \alpha_i P^\ast_i$. Let $v$ be a node with a coordinate $Q$.
In the following case, we can insert a convex conjugate centroid.
\begin{figure}[H]
 \begin{center}
  \includegraphics[width=120mm]{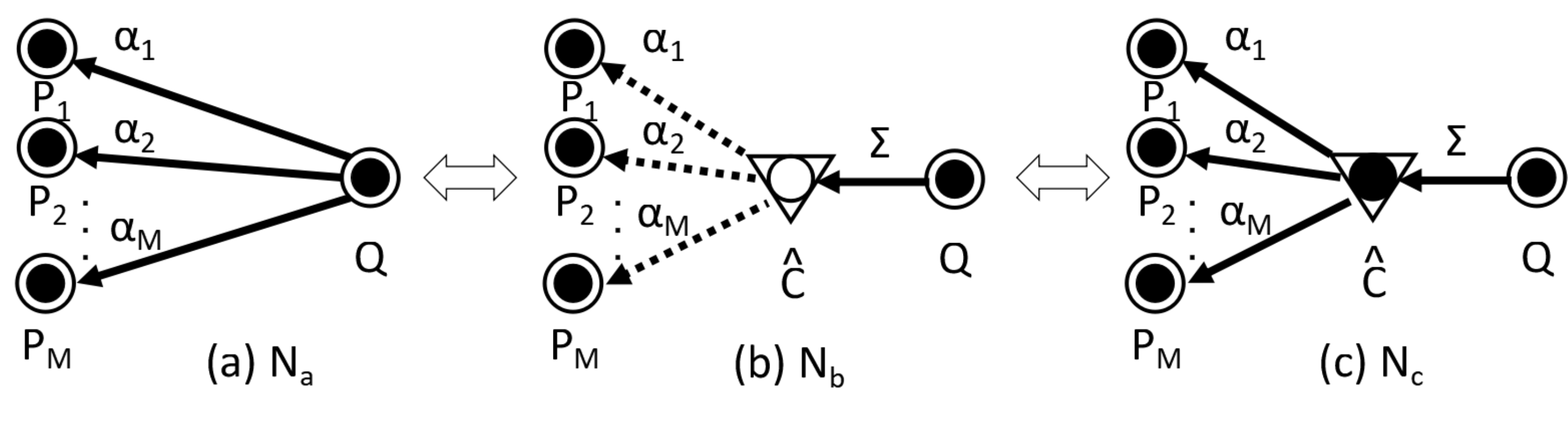}
 \end{center}
 \caption{Insertion rule 2.}
 \label{insertion_rule2}
\end{figure}
\noindent\textbf{Remark:} In $N_a$, $N_b$ and $N_c$, the state of nodes with coordinates $\{P_i\}$ and $Q$ are the same.\\
\noindent\textbf{Proof.}\\
We first prove Figure \ref{insertion_rule2} (a) $\rightarrow$ (b) in the case a node $Q$ is ON-state. By using the definition of centroids $\hat{C}^\ast\eqdef \frac{1}{\Sigma}\sum_i \alpha_i P^\ast_i$, we get 
\begin{align} 
\Phi_N(N_a)&= \sum_i\alpha_i\langle Q, P_i^\ast\rangle + \sum_i \alpha_i F(Q) + R(v)\\ \nonumber
&=\Sigma \langle Q, \hat{C}^\ast\rangle + \Sigma F(Q)+ R(v)\\
\Phi_N(N_b)&=\Sigma \langle Q, \hat{C}^\ast\rangle+ \Sigma F(Q)+ R(v),
\end{align}
where $R(v)$ denotes the sum of the values from nodes with coordinates $\{P_i\}$.
Then, the result follows. In the case a node $Q$ is OFF-state, we can prove in the same way.
We prove Figure \ref{insertion_rule2} (b) $\iff$ (c) in the same way as the insertion rule 1.

\subsection{Connection rule}
Let $C\eqdef \frac{1}{\alpha+\beta}(\alpha P + \beta Q)$ and $\hat{C}^\ast\eqdef \frac{1}{\alpha+\beta}(\alpha P^\ast + \beta Q^\ast)$.
In the following case, we can connect nodes on $P$ and $Q$ directly.
\begin{figure}[H]
 \begin{center}
  \includegraphics[width=80mm]{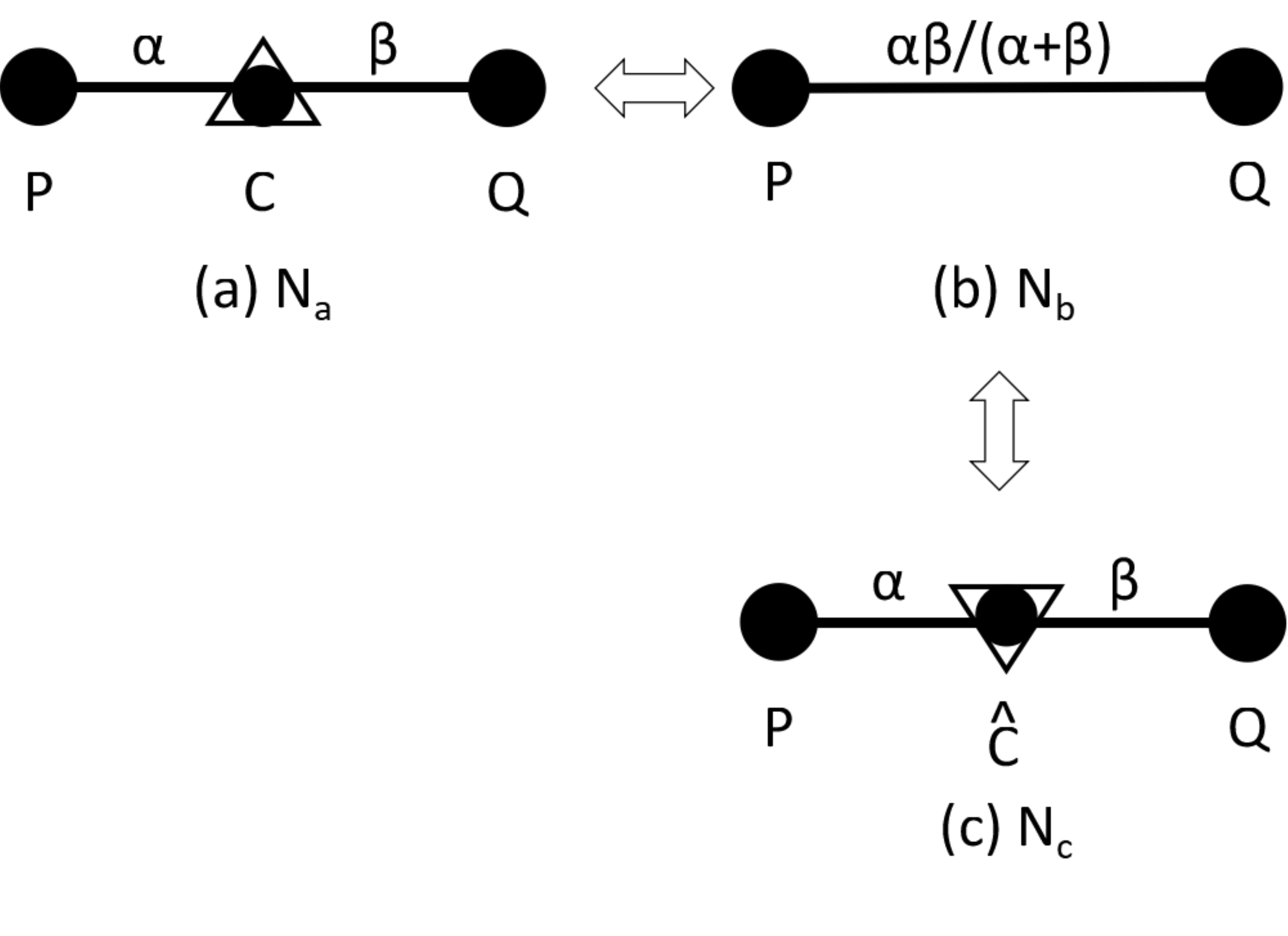}
 \end{center}
 \caption{Connection rule.}
 \label{connection_rule}
\end{figure}
\noindent\textbf{Proof.}\\
We prove Figure \ref{connection_rule} by using the deformation rules as follows.
\begin{figure}[H]
 \begin{center}
  \includegraphics[width=100mm]{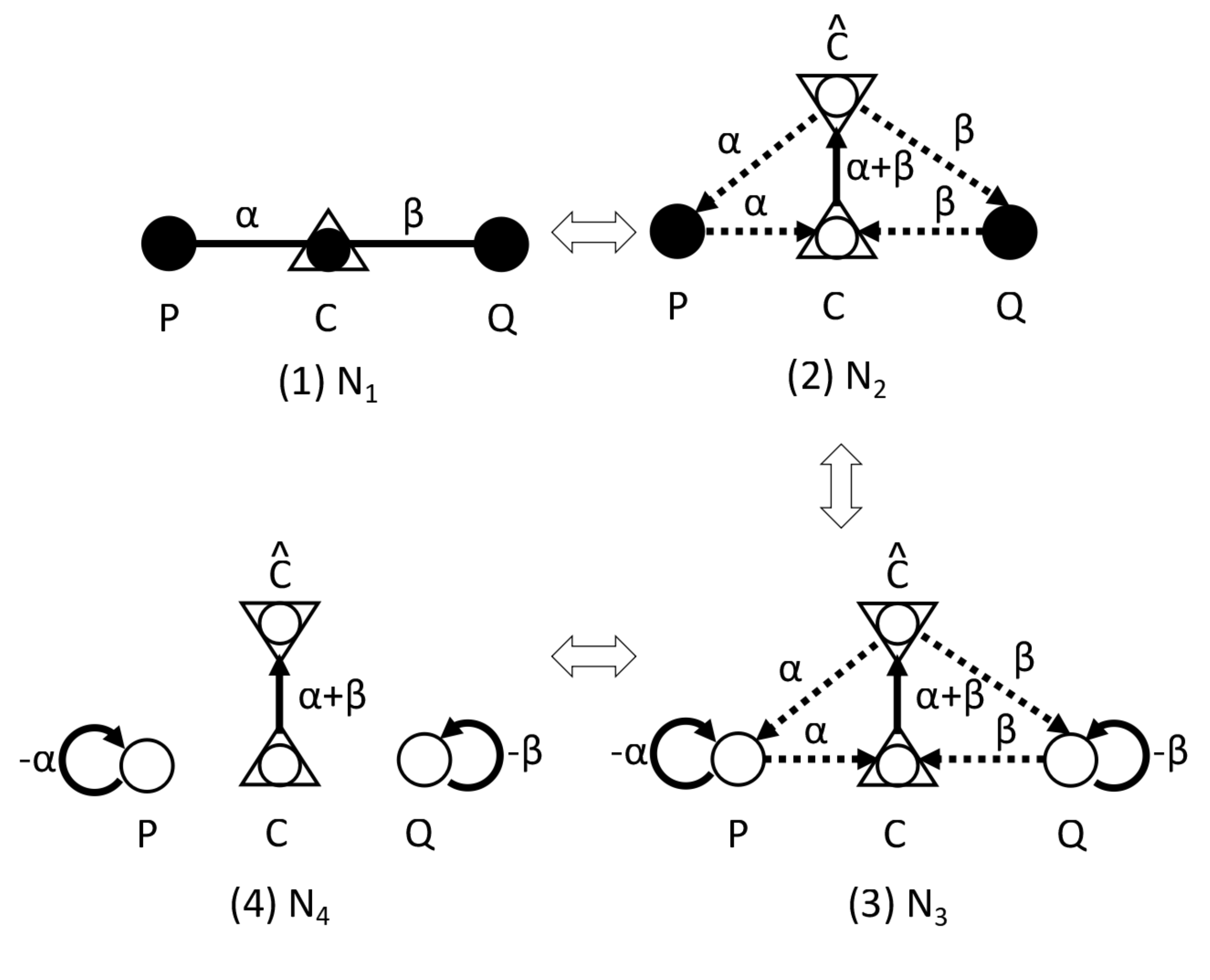}
 \end{center}
 \caption{Deformation of the network $N_a$.}
\label{deform_connection_rule}
\end{figure}
\begin{itemize}
\item (1)$\rightarrow$(2): We apply Figure \ref{insertion_rule2} (a)$\rightarrow$ (b) (insertion rule 2) for $Q=C$ and apply Figure \ref{insertion_rule1} (c) $\rightarrow$ (b) (insertion rule 1) for $Q=\hat{C}$.
\item (2)$\rightarrow$(3): We apply Figure \ref{onoff_rule1} (a)$\rightarrow$ (b) (ON-OFF rule 1) for $P$ and $Q$.
\item (3)$\rightarrow$(4): We apply Figure \ref{deletion_rule} (a) (deletion rule).
\end{itemize}
From this figure and the definitions of the centroid and the convex conjugate centroid, we get 
\begin{align}
\Phi_N(N_4)&=\alpha \langle P, P^\ast\rangle + \beta \langle Q, Q^\ast\rangle-(\alpha + \beta)\langle C,\hat{C}^\ast\rangle\\\nonumber
& =\frac{\alpha\beta}{\alpha+\beta}\bigl(\langle P, P^\ast\rangle + \langle Q, Q^\ast\rangle- \langle P, Q^\ast\rangle - \langle Q, P^\ast\rangle\bigr).
\end{align} 
On the other hand, by using (\ref{convex conjugate}), we get
\begin{align}
\Phi_N(N_b)&=\frac{\alpha\beta}{\alpha+\beta}\bigl(F(P) + F^\ast(P^\ast)- \langle P, Q^\ast\rangle - \langle Q, P^\ast\rangle+F(Q) + F^\ast(Q^\ast)\bigr)\\\nonumber
&=\frac{\alpha\beta}{\alpha+\beta}\bigl(\langle P, P^\ast\rangle + \langle Q, Q^\ast\rangle- \langle P, Q^\ast\rangle - \langle Q, P^\ast\rangle\bigr).
\end{align} 
Hence, we obtain $\Phi_N(N_a)=\Phi_N(N_4)=\Phi_N(N_b)$.
We can prove $\Phi_N(N_b)=\Phi_N(N_c)$ in the same way.

\section{Network representations of divergence functions}
We show some divergence network representations of divergence functions.
\subsection{Bregman network and symmetric Bregman network}
We have introduced the Bregman network in subsection \ref{Bregman_network}.
The Bregman network represents not only the Bregman divergence but also the canonical divergence of a dually flat space\cite{amari2010information}.\\
\textbf{Proof.}\\
The canonical divergence $D_C(P,Q)$ is defined as 
\begin{align}
D_C(P,Q)\eqdef \psi(\theta(P))+\phi(\eta(Q))- \sum_i \theta_i(P) \eta_i(Q),
\end{align}
where $\{\theta_i\}$ and $\{\eta_i\}$ denote affine coordinates which satisfy $\eta_i=\theta^\ast_i$ and $\psi(\theta)$ and $\phi(\eta)$ are strictly convex functions which satisfy $\phi=\psi^\ast$.
By replacing $F\rightarrow\psi$ and $P_i\rightarrow\theta_i(P)$ and using (\ref{eq_Bregman}), we get $\Phi(N_B)=\alpha D_C(P,Q)$.

Next, we introduce the symmetric Bregman network $N_{SB} $ as follows.
\begin{figure}[H]
 \begin{center}
  \includegraphics[width=30mm]{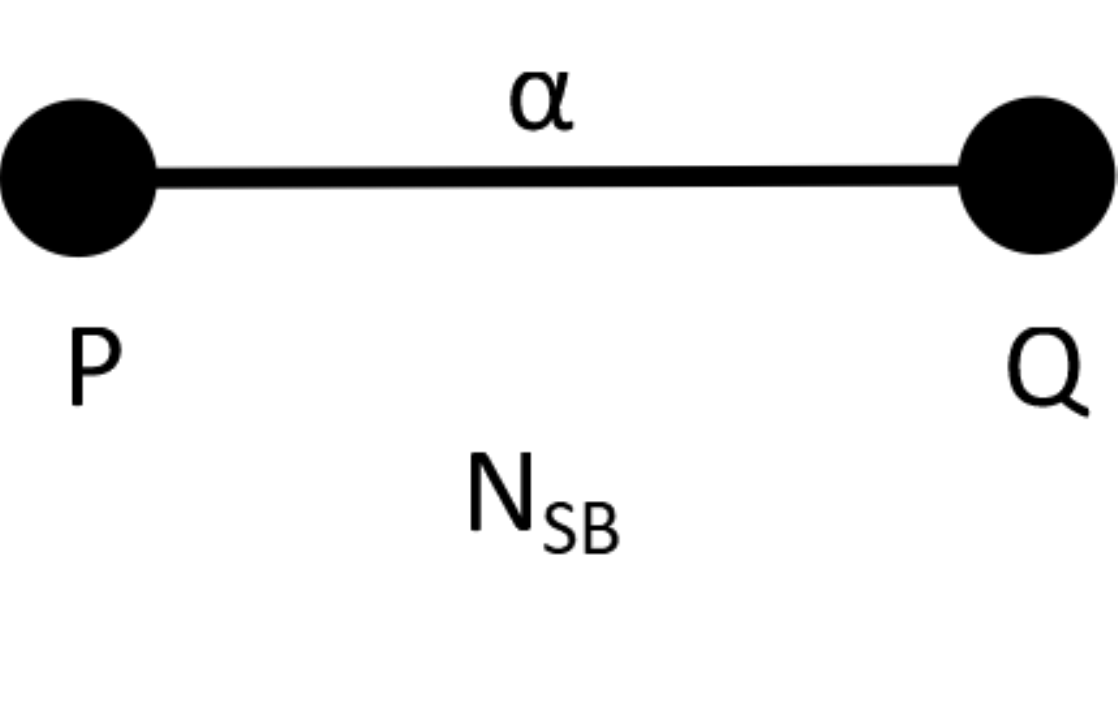}
 \end{center}
 \caption{symmetric Bregman network.}
\end{figure}
\begin{align}
\Phi(N_{SB})=\alpha B_{F,\mathrm{sym}}(P,Q),
\end{align}
where $B_{F,\mathrm{sym}}(P,Q)\eqdef B_{F}(P,Q) + B_{F}(Q,P)$ is the symmetric Bregman divergence\cite{nielsen2009sided}. \\
\textbf{Proof.}\\
By Figure \ref{def_lines} and Figure \ref{Bregman}, the result follows.

\subsection{Jensen network}
Let $\Sigma\eqdef \sum_i \alpha_i$ and $C\eqdef \frac{1}{\Sigma}\sum_i \alpha_i P_i$.
We introduce the Jensen network $N_J $ as follows.
\begin{figure}[H]
 \begin{center}
  \includegraphics[width=35mm]{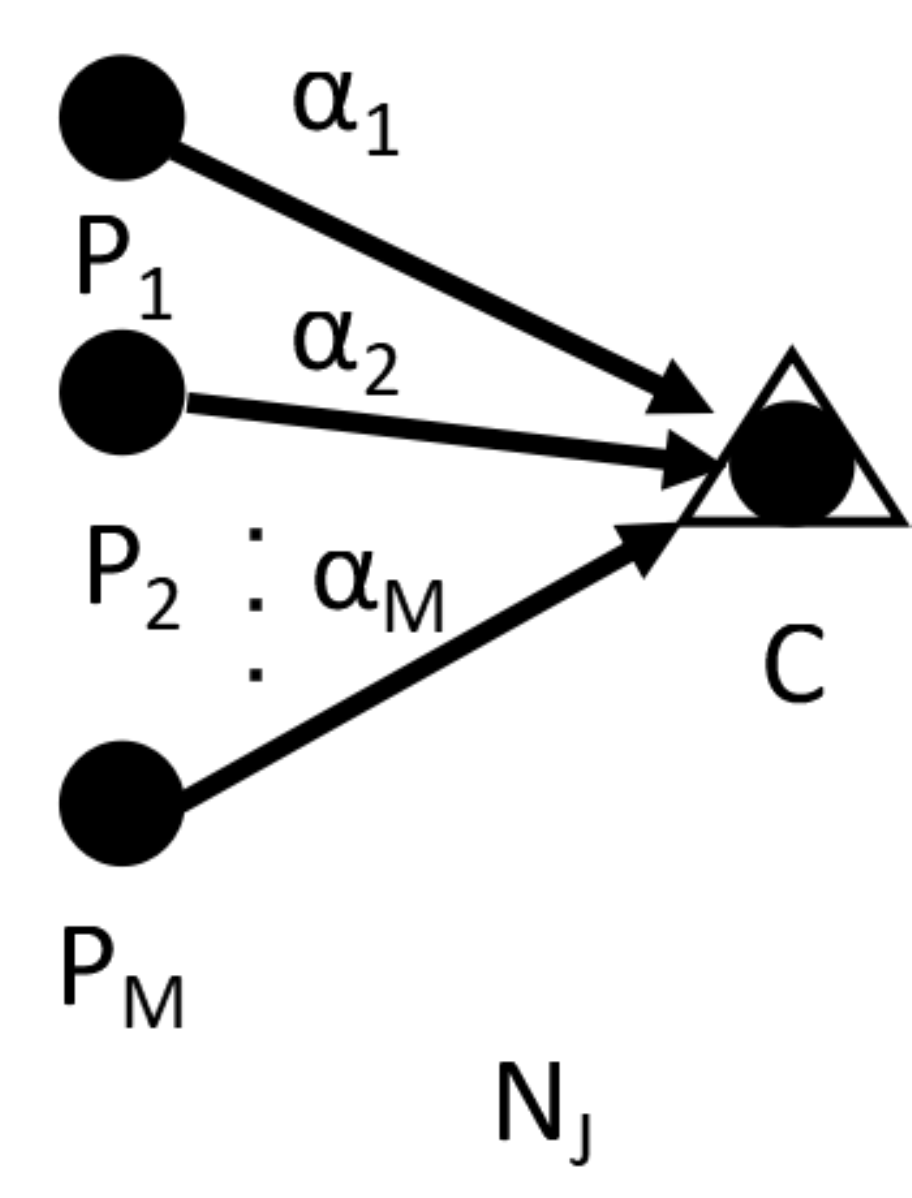}
 \end{center}
 \caption{Jensen network.}
 \label{Jensen}
\end{figure}
\begin{align}
\label{jensen_net}
\Phi(N_J)=\Sigma J_{F,\boldsymbol{\alpha}}(\bm{P}),
\end{align}
where $J_{F,\boldsymbol{\alpha}}(\bm{P})\eqdef \frac{1}{\Sigma}\sum_i \alpha_i F(P_i)-F(C)$, $\boldsymbol{\alpha}$ denotes $(\alpha_1, \alpha_2, \cdots \alpha_M)$ and $\bm{P}$ denotes $(P_1, P_2, \cdots P_M)$.

If $\alpha_i>0$ for all $i$, $J_{F,\boldsymbol{\alpha}}(\bm{P})$ is the Jensen divergence\cite{nielsen2017generalizing}. \\
\noindent\textbf{Proof.}\\
We deform the divergence network $N_J$ as follows.

\begin{figure}[H]
 \begin{center}
  \includegraphics[width=90mm]{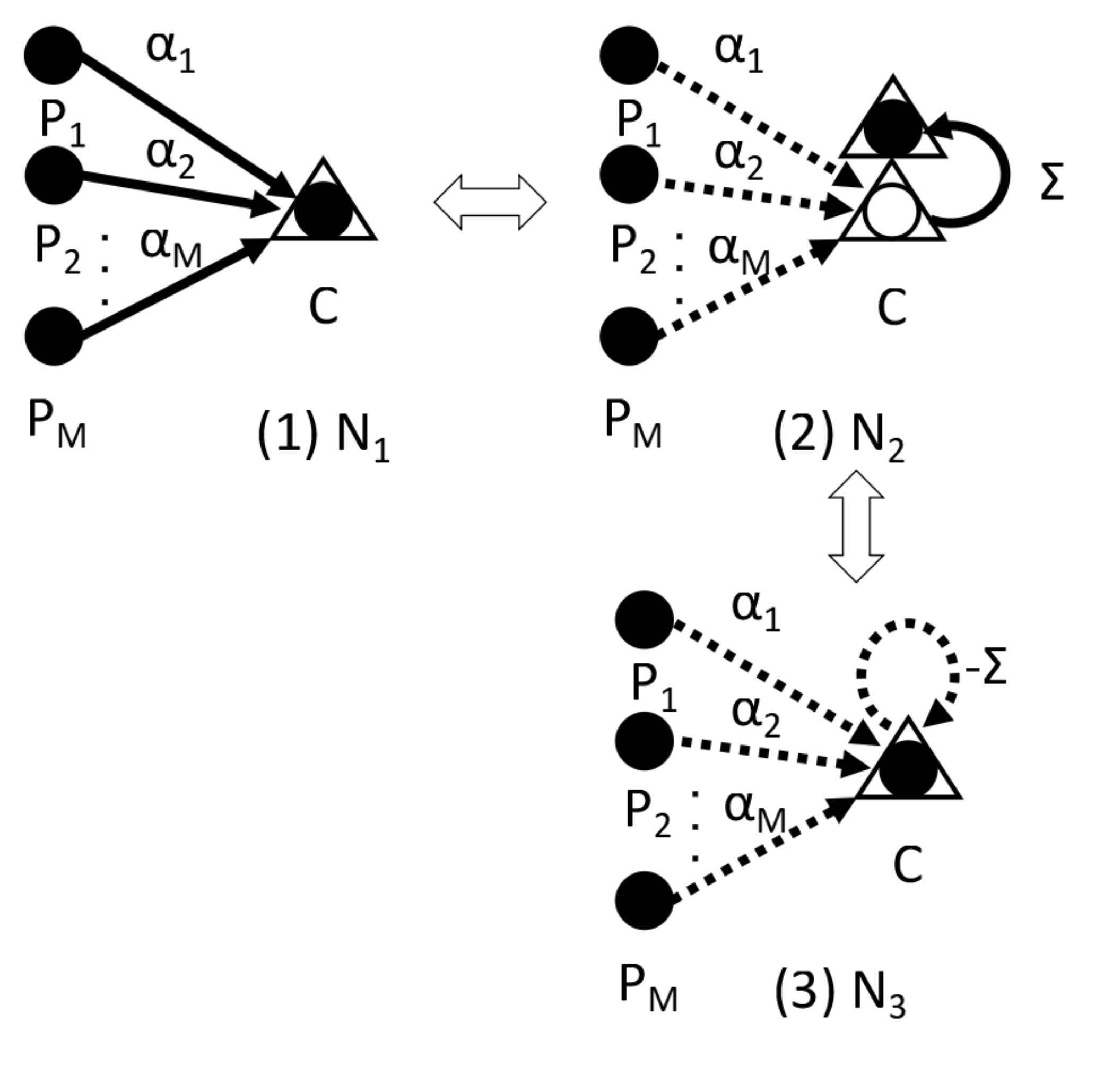}
 \end{center}
 \caption{Deformation of the network $N_J$.}
\end{figure}
\begin{itemize}
\item (1)$\rightarrow$(2): We apply Figure \ref{insertion_rule1} (a) $\rightarrow$ (b) (insertion rule 1) for $Q=C$.
\item (2)$\rightarrow$(3): We apply Figure \ref{onoff_rule2}  (a) $\rightarrow$ (b) (ON-OFF rule 2) for $C$ and apply Figure \ref{deletion_rule} (c) (deletion rule) for $C$.
\end{itemize}
\begin{align}
\Phi(N_3)&=\sum_i \alpha_i F(P_i)+\sum_i \alpha_i F^\ast(C^\ast)-\Sigma(F(C)+F^\ast(C^\ast))\\ \nonumber
&=\Sigma\bigl(\frac{1}{\Sigma}\sum_i \alpha_i F(P_i)-F(C)\bigr)
\end{align}
Hence, the result follows.

\subsection{Convex conjugate Jensen divergence}
Let $\Sigma\eqdef \sum_i \alpha_i$ and $\hat{C}^\ast\eqdef \frac{1}{\Sigma}\sum_i \alpha_i P^\ast_i$.
We introduce the convex conjugate Jensen network $N_{CJ} $ as follows.
\begin{figure}[H]
 \begin{center}
  \includegraphics[width=35mm]{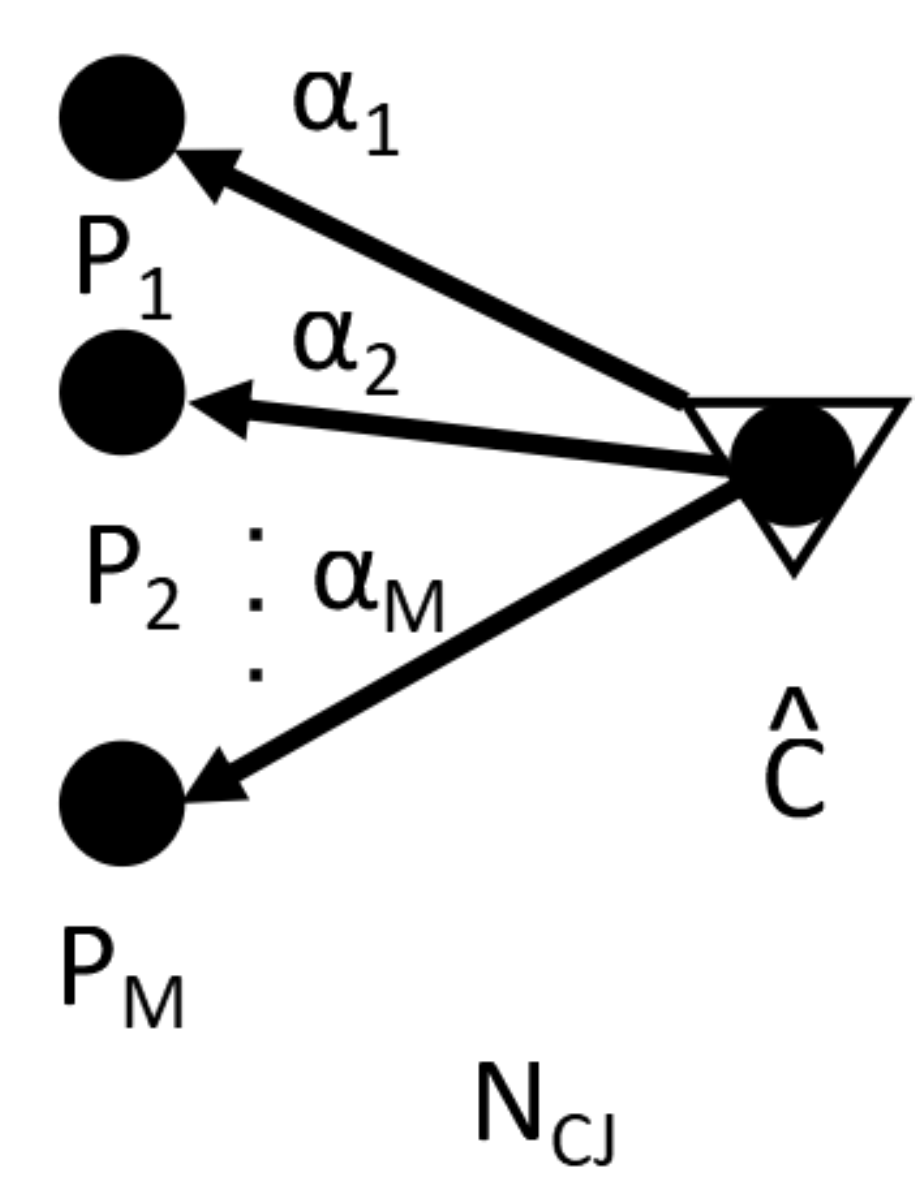}
 \end{center}
 \caption{Convex conjugate Jensen network.}
 \label{conjugateJensen}
\end{figure}
\begin{align}
\Phi(N_{CJ})=\Sigma J_{F^\ast,\boldsymbol{\alpha}}(\bm{P}^\ast),
\end{align}
where $J_{F^\ast,\boldsymbol{\alpha}}(\bm{P}^\ast)\eqdef \frac{1}{\Sigma}\sum_i \alpha_i F^\ast(P^\ast_i)-F^\ast(\hat{C}^\ast)$, $\boldsymbol{\alpha}$ denotes $(\alpha_1, \alpha_2, \cdots \alpha_M)$ and $\bm{P}^\ast$ denotes $(P^\ast_1, P^\ast_2, \cdots P^\ast_M)$.

If $\alpha_i>0$ for all $i$, $J_{F^\ast,\boldsymbol{\alpha}}(\bm{P}^\ast)$ is the convex conjugate Jensen divergence.
We can prove this equation in the same way as the Jensen network.

\subsection{$f$-network}
Let $F(1)=0$ and $\Sigma\eqdef\sum_i p_i=\sum_i q_i$.
We introduce the $f$-network $N_f$ as follows.
\begin{figure}[H]
 \begin{center}
  \includegraphics[width=40mm]{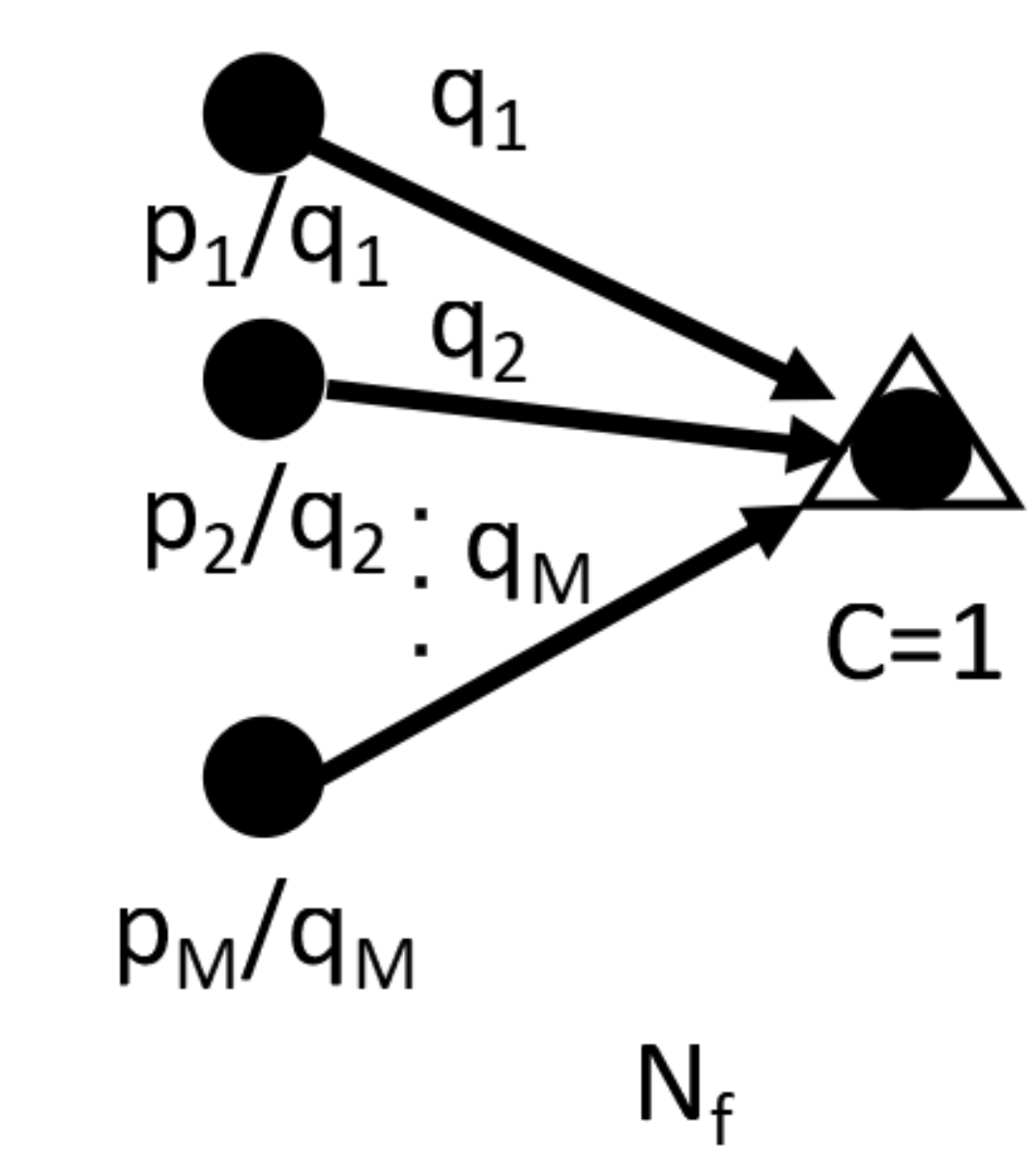}
 \end{center}
 \caption{$f$-network.}
 \label{f_divergence}
\end{figure}
\begin{align}
\Phi(N_f)=\Sigma D_f(P,Q), 
\end{align}
where $D_f(P,Q)\eqdef \frac{1}{\Sigma}\sum_i q_i F\bigl(\frac{p_i}{q_i}\bigr)$.
If $p_i,q_i>0$ for all $i$ and $\Sigma=1$,  $D_f(P,Q)$ is the $f$-divergence.\\
\noindent\textbf{Proof.}
We can write $D_f(P,Q)=\frac{1}{\Sigma}\sum_i q_i F\bigl(\frac{p_i}{q_i}\bigr)-F(1)$.
By replacing $P_i\rightarrow \frac{p_i}{q_i}$ and putting $\alpha_i=q_i$ in $J_{F,\boldsymbol{\alpha}}(\bm{P})=\frac{1}{\Sigma}\sum_i \alpha_i F(P_i)-F(C)$, we get $C=\frac{1}{\Sigma}\sum_i q_i\frac{p_i}{q_i}=1$ and the same equation as $D_f(P,Q)$.
Then, the result follows.
The network topology of the Jensen network and $f$-network are the same.

\section{Application examples of the divergence networks}
Let $\sum_i$ denotes $\sum_{i=1}^M$ and $\Sigma=\sum_i \alpha_i$.
We show some results by using the divergence networks.
\subsection{Results about relations among different divergences}
By comparing Figure \ref{Bregman} and Figure \ref{Jensen} and using the network function Property 1, we get
\begin{align}
\frac{1}{\Sigma}\sum_i \alpha_i B_F(P_i, C)=J_{F,\boldsymbol{\alpha}}(\bm{P}).
\end{align}
By comparing Figure \ref{Bregman} and Figure \ref{conjugateJensen} and using the network function Property 1, we get
\begin{align}
\frac{1}{\Sigma}\sum_i \alpha_i B_F(\hat{C},P_i)=J_{F^\ast,\boldsymbol{\alpha}}(\bm{P}^\ast).
\end{align}
By putting $Q=C$ in Figure \ref{insertion_rule2} (insertion rule 2), we get
\begin{align}
\sum_i \alpha_i B_F(C,P_i)=\sum_i \alpha_i B_F(\hat{C},P_i)+\Sigma B_F(C,\hat{C})=\Sigma \bigl(J_{F^\ast,\boldsymbol{\alpha}}(\bm{P}^\ast)+B_F(C,\hat{C})\bigr).
\end{align}
Adding $\sum_i \alpha_i B_F(P_i,C)=\Sigma J_{F,\boldsymbol{\alpha}}(\bm{P})$ to both sides, we get
\begin{align}
\label{Bregman_Jensen}
\frac{1}{\Sigma}\sum_i \alpha_i B_{F,\mathrm{sym}}(P_i,C)=J_{F,\boldsymbol{\alpha}}(\bm{P})+J_{F^\ast,\boldsymbol{\alpha}}(\bm{P}^\ast)+B_F(C,\hat{C}).
\end{align}
We can deform the divergence network of LHS of this equation in the same way as Figure \ref{deform_connection_rule}.
\begin{figure}[H]
 \begin{center}
  \includegraphics[width=90mm]{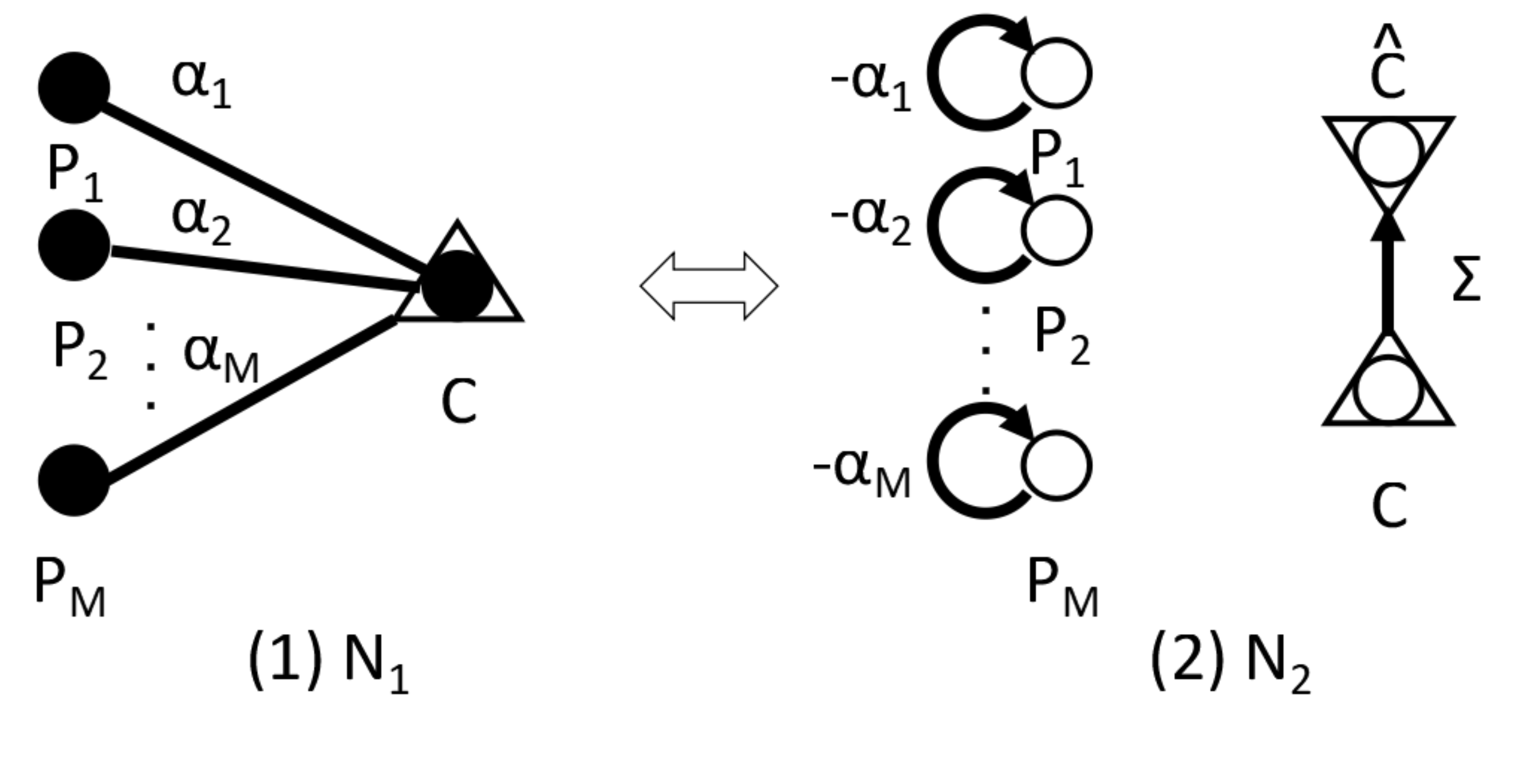}
 \end{center}
 \caption{Deformation of the divergence network.}
\end{figure}
From this figure, we get
\begin{align}
\sum_i \alpha_i B_{F,\mathrm{sym}}(P_i,C)=\sum_i\alpha_i \langle P_i,P_i^\ast\rangle -\Sigma \langle C,\hat{C}^\ast\rangle.
\end{align}
By combining (\ref{Bregman_Jensen}), we get 
\begin{align}
\label{Bregman_Jensen2}
\frac{1}{\Sigma}\sum_i\alpha_i \langle P_i,P_i^\ast\rangle - \langle C,\hat{C}^\ast\rangle=J_{F,\boldsymbol{\alpha}}(\bm{P})+J_{F^\ast,\boldsymbol{\alpha}}(\bm{P}^\ast)+ B_F(C,\hat{C}).
\end{align}
If $M=2$, (\ref{Bregman_Jensen}) is written as
\begin{align}
\frac{1}{\alpha+\beta}\bigl(\alpha B_{F,\mathrm{sym}}(P,C)+\beta B_{F,\mathrm{sym}}(Q,C)\bigr)=J_{F,(\alpha,\beta)}(P,Q)+J_{F^\ast,(\alpha,\beta)}(P^\ast, Q^\ast)+B_F(C,\hat{C}).
\end{align}
From Figure \ref{connection_rule} (connection rule), we get
\begin{align}
\frac{\alpha\beta}{{(\alpha+\beta)}^2}B_{F,\mathrm{sym}}(P,Q)=J_{F,(\alpha,\beta)}(P,Q)+J_{F^\ast,(\alpha,\beta)}(P^\ast, Q^\ast)+B_F(C,\hat{C}).
\end{align}

For example, when $F(x)=\sum_{\mu=1}^d x_\mu\log x_\mu$ and $\alpha +\beta=1$, $B_F(P,Q)$, $B_{F,\mathrm{sym}}(P,Q)$, $\frac{1}{\alpha\beta}J_{F,(\alpha,\beta)}(P,Q)$ and $\frac{1}{\alpha\beta}J_{F^\ast,(\alpha,\beta)}(P^\ast, Q^\ast)$ are equal to the Kullback-Leibler divergence\cite{kullback1997information}, the Jeffrey's J-divergence\cite{jeffreys1946invariant}, the scaled skew Jensen-Shannon divergence\cite{nielsen2011burbea} and the Amari's $\alpha$-divergence\cite{cichocki2010families} respectively (see \cite{nishiyama2018sum}).

By replacing $P_i\rightarrow \frac{p_i}{q_i}$ and putting $\alpha_i=q_i$ in (\ref{Bregman_Jensen2}) and using $C=1$ and $\hat{C}^\ast=\sum_i q_i {\bigl(\frac{p_i}{q_i}\bigr)}^\ast=\sum_i q_i F'\bigl(\frac{p_i}{q_i}\bigr)$, we get
\begin{align}
\label{eq_f_divergence}
\frac{1}{\Sigma}\sum_i F'\bigl(\frac{p_i}{q_i}\bigr)(p_i-q_i)=D_f(P,Q)+\hat{D}_f(P,Q)+B_F(1,\hat{C}), 
\end{align}
where $\hat{D}_f(P,Q)\eqdef=\frac{1}{\Sigma}\sum_i q_i F^\ast\bigl(F'\bigl(\frac{p_i}{q_i}\bigr)\bigr) -F^\ast(\hat{C}^\ast)$.

For example, when $F(x)=-\log x$ and $\Sigma=1$, $B_F(P,Q)$ and $D_f(P,Q)$ are equal to the Itakura-Saito divergence\cite{itakura1968analysis} and the reverse Kullback-Leibler divergence.
The LHS of (\ref{eq_f_divergence}) is equal to the Neyman chi-square divergence\cite{cichocki2010families} (see \cite{nishiyama2018sum}).
\subsection{Results about geometrical properties of divergence functions}
We show the parallelogram law for the symmetric Bregman divergence.
Let $P_i+R_i=Q_i+S_i$ for all $i$ or $P^\ast_i+R^\ast_i=Q^\ast_i+S^\ast_i$ for all $i$.
Then, 
\begin{align}
\label{eq_parallelogram}
B_{F,\mathrm{sym}}(P,Q)+B_{F,\mathrm{sym}}(Q,R)+B_{F,\mathrm{sym}}(R,S)+B_{F,\mathrm{sym}}(S,P)\\ \nonumber
=B_{F,\mathrm{sym}}(P,R)+B_{F,\mathrm{sym}}(Q,S)
\end{align}
holds.
In the case $P_i+R_i=Q_i+S_i$, we prove this equation by deforming the divergence network as follows. In the case $P^\ast_i+R^\ast_i=Q^\ast_i+S^\ast_i$, we can prove in the same way.
\begin{figure}[H]
 \begin{center}
  \includegraphics[width=90mm]{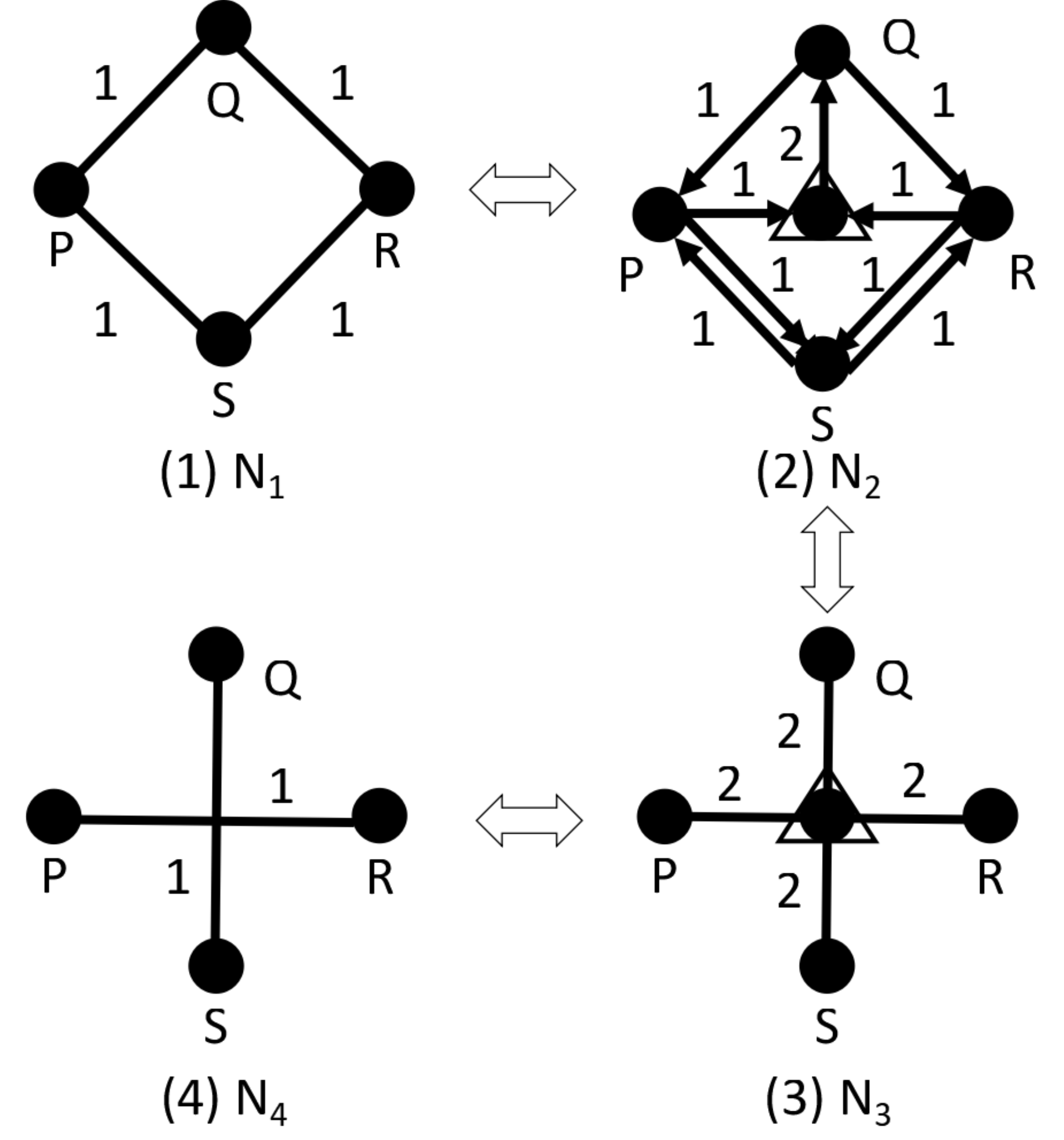}
 \end{center}
 \caption{Deformation of the divergence network.}
\end{figure}
\begin{itemize}
\item (1)$\rightarrow$(2): We apply Figure \ref{insertion_rule1} (a) $\rightarrow$ (c) (insertion rule 1) for triangle $PQR$. $C$ is a centroid of $P$ and $R$.
\item (2)$\rightarrow$(3): By assumption, centroids of $P$ and $R$, $Q$ and $S$ and rectangle $PQRS$ are the same. Hence, we can apply Figure \ref{insertion_rule1} (a) $\rightarrow$ (c) (insertion rule 1) for triangle $QRS$, $RSP$ and $SPQ$ in the same way. 
\item (3)$\rightarrow$(4): We apply Figure \ref{connection_rule} (a) $\rightarrow$ (b) (connection rule).
\end{itemize}
As described above, we can grasp relations among divergence functions and geometric properties of them intuitively.
When $F(x)=\frac{1}{2}\sum_{\mu=1}^d x_\mu^2$, $B_{F,\mathrm{sym}}(P,Q)$ is equal to the squared Euclidean distance and (\ref{eq_parallelogram}) is consistent with the parallelogram law in the Euclidean space.

\section{Conclusion}
We have introduced a new graphical calculation method called ``divergence network'' for divergence functions and have introduced the network function which gives the value of the divergence network.

Then, we have shown properties of the network function and the deformation rules of the divergence networks. 

Finally, we have shown examples of the network representations of divergence functions and have shown some application examples of the divergence networks.

Through application examples, we have concluded that it is easy to intuitively grasp relations among divergence functions and geometric meaning of the results by using the divergence networks.

Studies on relation with graph theory or network theory are future works.
\bibliography{reference_v1}

\begin{thebibliography}{10}

\bibitem{ali1966general}
Syed~Mumtaz Ali and Samuel~D Silvey.
\newblock A general class of coefficients of divergence of one distribution
  from another.
\newblock {\em Journal of the Royal Statistical Society. Series B
  (Methodological)}, pages 131--142, 1966.

\bibitem{amari2010information}
Shun-ichi Amari and Andrzej Cichocki.
\newblock Information geometry of divergence functions.
\newblock {\em Bulletin of the Polish Academy of Sciences: Technical Sciences},
  58(1):183--195, 2010.

\bibitem{bregman1967relaxation}
Lev~M Bregman.
\newblock The relaxation method of finding the common point of convex sets and
  its application to the solution of problems in convex programming.
\newblock {\em USSR computational mathematics and mathematical physics},
  7(3):200--217, 1967.

\bibitem{burbea1980convexity}
Jacob Burbea and C~Radhakrishna Rao.
\newblock On the convexity of some divergence measures based on entropy
  functions.
\newblock Technical report, PITTSBURGH UNIV PA INST FOR STATISTICS AND
  APPLICATIONS, 1980.

\bibitem{cichocki2010families}
Andrzej Cichocki and Shun-ichi Amari.
\newblock Families of alpha-beta-and gamma-divergences: Flexible and robust
  measures of similarities.
\newblock {\em Entropy}, 12(6):1532--1568, 2010.

\bibitem{csiszar1967information}
Imre Csisz{\'a}r.
\newblock Information-type measures of difference of probability distributions
  and indirect observation.
\newblock {\em studia scientiarum Mathematicarum Hungarica}, 2:229--318, 1967.

\bibitem{itakura1968analysis}
Fumitada Itakura.
\newblock Analysis synthesis telephony based on the maximum likelihood method.
\newblock In {\em The 6th international congress on acoustics, 1968}, pages
  280--292, 1968.

\bibitem{jeffreys1946invariant}
Harold Jeffreys.
\newblock An invariant form for the prior probability in estimation problems.
\newblock {\em Proc. R. Soc. Lond. A}, 186(1007):453--461, 1946.

\bibitem{kullback1997information}
Solomon Kullback.
\newblock {\em Information theory and statistics}.
\newblock Courier Corporation, 1997.

\bibitem{nielsen2011burbea}
Frank Nielsen and Sylvain Boltz.
\newblock The burbea-rao and bhattacharyya centroids.
\newblock {\em IEEE Transactions on Information Theory}, 57(8):5455--5466,
  2011.

\bibitem{nielsen2009sided}
Frank Nielsen and Richard Nock.
\newblock Sided and symmetrized bregman centroids.
\newblock {\em IEEE transactions on Information Theory}, 55(6):2882--2904,
  2009.

\bibitem{nielsen2017generalizing}
Frank Nielsen and Richard Nock.
\newblock Generalizing jensen and bregman divergences with comparative
  convexity and the statistical bhattacharyya distances with comparable means.
\newblock {\em arXiv preprint arXiv:1702.04877}, 2017.

\bibitem{nishiyama2018generalized}
Tomohiro Nishiyama.
\newblock Generalized bregman and jensen divergences which include some
  f-divergences.
\newblock {\em arXiv preprint arXiv:1808.06148}, 2018.

\bibitem{nishiyama2018sum}
Tomohiro Nishiyama.
\newblock Sum decomposition of divergence into three divergences.
\newblock {\em arXiv preprint arXiv:1810.01720}, 2018.

\end{thebibliography}
\end{document}